\documentclass[conference]{IEEEtran}
\IEEEoverridecommandlockouts

\usepackage[utf8]{inputenc}
\usepackage[T1]{fontenc}
\usepackage{graphicx}            
\usepackage[dvipsnames]{xcolor}   
\usepackage{amsmath,amssymb,bm}  
\usepackage{cite}                
\usepackage{hyperref}            
\usepackage{cleveref} 
\usepackage{colortbl}        
\usepackage{array,tabularx,booktabs,threeparttable,longtable,multirow,makecell}
\usepackage[ruled,vlined]{algorithm2e}

\usepackage{caption}     
\usepackage{subcaption}  
\usepackage{pifont}      
\newcommand{\cmark}{\textcolor{green!70!black}{\ding{51}}}
\newcommand{\xmark}{\textcolor{red!80!black}{\ding{55}}}

\begin{document}

\title{Efficiency Bottlenecks of Convolutional Kolmogorov-Arnold Networks: A Comprehensive Scrutiny with ImageNet, AlexNet, LeNet and Tabular Classification}

\author{\IEEEauthorblockN{Ashim Dahal\textsuperscript{1}, Saydul Akbar Murad\textsuperscript{1}, and Nick Rahimi\textsuperscript{1}}
\IEEEauthorblockA{\textsuperscript{1}\textit{School of Computing Sciences and Computer Engineering}\\
\textit{University of Southern Mississippi}\\
Hattiesburg, MS, 39406, USA\\
\{ashim.dahal, saydulakbar.murad, nick.rahimi\}@usm.edu}
}

\maketitle

\begin{abstract}
Algorithmic level developments like Convolutional Neural Networks, transformers, attention mechanism, Retrieval Augmented Generation and so on have changed Artificial Intelligence. Recent such development was observed by Kolmogorov-Arnold Networks that suggested to challenge the fundamental concept of a Neural Network, thus change Multilayer Perceptron, and Convolutional Neural Networks. They received a good reception in terms of scientific modeling, yet had some drawbacks in terms of efficiency. In this paper, we train Convolutional Kolmogorov Arnold Networks (CKANs) with the ImageNet-1k dataset with 1.3 million images, MNIST dataset with 60k images and a tabular biological science related MoA dataset and test the promise of CKANs in terms of FLOPS, Inference Time, number of trainable parameters and training time against the accuracy, precision, recall and f-1 score they produce against the standard industry practice on CNN models. We show that the CKANs perform fair yet slower than CNNs in small size dataset like MoA and MNIST but are not nearly comparable as the dataset gets larger and more complex like the ImageNet. The code implementation of this paper can be found on the link: \href{https://github.com/ashimdahal/Study-of-Convolutional-Kolmogorov-Arnold-networks}{https://github.com/ashimdahal/Study-of-Convolutional-Kolmogorov-Arnold-networks}
\end{abstract}


\section{Introduction}
Deep learning and Neural Networks are fundamental to  Artificial Intelligence research \cite{sejnowski2018deep, sejnowski2020unreasonable}. Recent advancements in Computer Vision, Natural Language Processing or Multimodal AI, all emerge from different sets of algorithms like Stable Diffusions \cite{ho2020denoising}, Convolutional Neural Networks (CNNs) \cite{lecun1989backprop}, Long Short Term Memory (LSTMs) \cite{hochreiter1997long}, Retrieval Augmented Generation (RAG) \cite{lewis2020retrieval}, Transformers \cite{vaswani2017attention} and so on. At the heart of all these deep learning algorithms are the fundamental idea of a Neural Network; an algorithm that establishes a non-linear relationship between the independent and dependent variable making use of weights, biases and activation function. Such networks have advanced the applied research fields of Biomedical Imaging, text and image generation, physical modeling, remote sensing, polymer science, etc. 

Recently, however, there have been new advancements in the field of Deep Learning that challenges to change the fundamental way researchers think about a Neural Network by challenging the traditionally utilized Multi Layer Perceptrons (MLPs) with Kolmogorov-Arnold Networks (KANs) \cite{liu2024kan}. The high level intuition of KANs is instead of training the weights and biases for any given representation and then passing it through the activation function, the entire activation function itself is trained. These functions are called B splines and are discussed in greater detail in section \ref{lit_review}. Proponents of KAN claim that KAN can yield similar if not better performance in terms of accuracy, precision and recall as compared to MLP with much less numbers of parameters in them while adversary research claim the exact opposite \cite{kan1,kan2,kan3,kan4,kan5}. This has left researchers divided on whether to adopt KANs in their current approach for reliable and tangible results.

Specific to the scope of this paper, we train and test Convolutional Kolmorogov Arnold Networks (CKANs)\cite{bodner2024convolutional} on the traditionally CNN focused tasks. We adopt comparable sizes of AlexNet\cite{alexnet}, LeNet \cite{lecun1989backprop} and 1-D tabular CNN \cite{guo2020kaggle} using CKAN implementation and compare the fidelity of the results directly to the CNN counterparts. Mainly, we focus on the Imagenet dataset \cite{deng2009imagenet}, MNIST dataset \cite{lecun1998mnist} and the Mechanisms of Action (MOA) \cite{paik2020moa} dataset to compare the three models in their own dominant regions. 

Our findings lay a strong foundation in adoption of CKANs not only in terms of fidelity of the results but also on the adaptability of CKANs in terms of training and testing efficiency. Mainly this paper brings the following novel contributions to the field of Convolutional Neural Networks:

\begin{itemize}
    \item CKAN can provide comparable results to CNN on small dataset like MNIST
    \item In a larger, modern-day medium, sized dataset like the ImageNet, CKAN cannot extrapolate or replicate its results
    \item Further refinement of the CKAN algorithm is required (which may not be possible with present computing) in order to make them a contender in the computer vision space
    \item CKANs are comparatively better performing in sciences and tabular CNN implementation as compared to computer vision task, although they still lag behind the SoTA CNN models
\end{itemize}

The rest of the paper is organized in the following order: in section \ref{lit_review}, we review the current literature available on CKANs and summarize them shortly, in section \ref{Section:methodology} we explain our research methodology and the reasoning behind our choices for the experiment. In section \ref{section:results}, we show the findings of our research objectives and in section \ref{section:conclusion}, we conclude the paper and discuss of potential future research in the field.

\section{Literature Review}\label{lit_review}

Liu et al \cite{liu2024kan} proposed Kolmogorov-Arnold Networks (KANs) based on the Kolmogorov-Arnold representation theorem. Established by Vladimir Arnold and Andrey Kolmogorov, the theorem states that if $f$ is a multivariate continuous function on a bounded domain, then $f$ can be written as a composition of multiple single variable functions summed together. Mathematically, $f:[0,1]^n \rightarrow \mathbb{R}$,

\begin{equation}
f(X) = f(x_1,x_2,...,x_n) = \sum_{q=1}^{2n+1} \Phi_q(\sum_{p=1}^n \phi_{qp}(x_p)),
\label{eq:kan}
\end{equation}

Where $\phi_{qp}: [0,1] \rightarrow \mathbb{R}$ and $\Phi_q: \mathbb{R} \rightarrow \mathbb{R}$

Kolmogorov-Arnold representation shows that the only true multivariate function is addition, since every other function can be written as a sum of multiple univariate functions \cite{liu2024kan}. The problem that arises with such a simple representation of a network with just two layers of non linearity and $(2n + 1)$ number of parameters in the hidden layer is that these networks may not be easily trainable in practice \cite{poggio2020theoretical, girosi1989representation}. Hence, the authors modified the network in two fundamental ways, first by adding arbitrary width and depths; i.e, ignoring $(2n + 1)$ in equation~\ref{eq:kan}, second by choosing the B-spline \cite{deboor1986basic} curve function to represent the univariate function $\Phi$. The authors then arrive with the following representation for a KAN architecture with $L$ layers where the $l^{th}$ layer $\Phi_{l}$ have shape $(n_{l + 1},n_{l})$:

\begin{equation}
KAN(x) = \Phi_{L-1}.\Phi_{L-2}...\Phi_{1}.\Phi_{0}.x
\label{eq:kan_architecture}
\end{equation}

The authors claimed KANs to be promising alternatives to MLPs, a standard form of representing Neural Networks in the field of Deep Learning. They showed proofs on a few datasets that made KANs outperform MLPs in terms of accuracy and interpretability on small scale AI alongside science tasks. The question to the efficacy of KANs as a viable option to MLPs come with the type of the dataset that the authors choose to use. For their demonstration purposes they choose to use toy datasets which could fail to encompass the huge amount of complexities and variability that come with real world complex datasets. The authors especially highlight the effectiveness of KANs in mathematical modeling tasks with some real world problems while sticking to toy datasets for other tasks.

The reasoning behind this choice is revealed by Yu et al \cite{yu2024kan} who propose that comparable MLPs are only inferior to KANs in symbolic formula representation tasks while being superior in other machine learning, computer vision, natural language processing and audio processing tasks. The authors deduced the number of parameters required for any particular KAN or MLP layer by using the formula in eq \ref{eq:kan_params}.

\begin{equation}
parameters_{KAN} = (d_{in} * d_{out}) * (G + K + 3) + d_{out}
\label{eq:kan_params}
\end{equation}

\begin{equation}
parameters_{MLP} = (d_{in}*d_{out}) + d_{out}
\label{eq:mlp_params}
\end{equation}

Where $d_{in}$ and $d_{out}$ are the dimensions of the input and output layer in the network, $G$ is the number of spline intervals and $K$ represents the order of the polynomial of the B-spline function.

From direct comparisons of the equations~\ref{eq:kan_params} and~\ref{eq:mlp_params}, it can be observed that if the number of input and output dimensions for the network layer were to be kept the same then the number of trainable parameters for KANs would grow much higher. For fairer comparisons, it thus becomes necessary to account for the higher number of trainable parameters in KANs to then reduce the input and output dimension for each layer in the KAN accordingly to maintain a fair playing ground with MLP.

Specific to the interest of this paper, in the department of computer vision, Yu et al \cite{yu2024kan} trained KAN and MLP with 8 standard computer vision datasets: MNIST \cite{lecun1998mnist}, EMNIST-Balanced \cite{cohen2017emnist}, EMNIST-Letters, FMNIST \cite{xiao2017fashion}, KMNIST \cite{clanuwat2018deep}, CIFAR10 \cite{krizhevsky2009learning}, CIFAR100 and SVHN \cite{netzer2011reading}. The hidden layer widths for MLP were 32, 64, 128, 256, 512 or 1024 while for KAN were 2, 4, 8 or 16 with respective B-spline grid 3, 5, 10 or 20 and B-spline degrees were 2, 3 or 5. Upon comparisons of these two models in the given 8 datasets, the authors concluded that KANs even with the highest number of parameters cannot surpass the accuracy of an MLP with the lowest number of parameters; while sometimes performing even worse than having a lower number of parameters in its own category.

Although authors from \cite{yu2024kan} provide concluding evidence on the superiority of MLPs over KANs in the domain of computer vision, the authors fail to accommodate for the new advancement on CNNs based on KANs proposed by Bodner et al \cite{bodner2024convolutional}. The authors from \cite{bodner2024convolutional} propose a new method of making CNN layers by decomposing the kernels of the CNN architecture as a KAN representation presented in \cite{liu2024kan}; i.e instead of learning the weights and biases for each of the filters in the CNN layer, the authors propose to learn the B-spline activation function like a KAN. The authors describe an image as follows.

\begin{equation}
image_{i,j} = a_{i,j}, 
\label{eq:image_def}
\end{equation}
 where  $a_{i,j}$  is a matrix of pixel values of size $m_1, m_2$  $(height, width)$

Then a KAN based kernel $K$ is described as $K \in \mathbb{R}^{n_2,n_1}$ which is a collection of B-splines defined as follows.

\begin{table*}[htbp]
\renewcommand{\arraystretch}{1.25}
\caption{Recent CKAN literature and their experimental coverage.}
\label{tab:research_summary}
\centering
\begin{threeparttable}
\begin{tabular}{l | cccc | ccc | ccc | c}
\hline
\multirow{2}{*}{\textbf{Paper}} &
\multicolumn{4}{c|}{\textbf{Models}} &
\multicolumn{3}{c|}{\textbf{Datasets}} &
\multicolumn{3}{c|}{\textbf{Performance Metrics}} &
\textbf{Ablation}\\
\cline{2-5}\cline{6-8}\cline{9-11}
& CKAN & AlexNet & LeNet & \makecell{Tabular\\CNN} &
MNIST & ImageNet & MoA &
FLOPs & Time & Params & \textbf{Sweeps} \\
\hline
Liu et al.\,\cite{liu2024kan}                & \xmark & \xmark & \xmark & \xmark & \xmark & \xmark & \xmark & \xmark & \cmark & \cmark & – \\
Yu et al.\,\cite{yu2024kan}                  & \xmark & \xmark & \xmark & \xmark & \cmark & \xmark & \xmark & \cmark & \xmark & \cmark & – \\
Bodner et al.\,\cite{bodner2024convolutional}& \cmark & \xmark & \cmark & \xmark & \cmark & \xmark & \xmark & \xmark & \cmark & \cmark & – \\
Guo\,\cite{guo2020kaggle}                    & \xmark & \xmark & \xmark & \cmark & \xmark & \xmark & \cmark & \xmark & \xmark & \xmark & – \\
Cang et al.\,\cite{ckan_potential}           & \cmark & \xmark & \xmark & \xmark & \xmark & \xmark & \xmark & \xmark & \xmark & \cmark & 4 \\
Cacciatore et al.\,\cite{kan_prelim}         & \cmark & \xmark & \xmark & \xmark & \cmark & \xmark & \xmark & \cmark & \cmark & \cmark & – \\
Jamali et al.\,\cite{kan_rs}                 & \cmark & \xmark & \xmark & \cmark & \xmark & \xmark & \xmark & \xmark & \xmark & \cmark & – \\
Kilani\,\cite{kan_key}                       & \cmark & \xmark & \xmark & \xmark & \xmark & \xmark & \xmark & \xmark & \xmark & \cmark & – \\ 
Cheon et al.\,\cite{cheon2024demonstrating}     & \xmark & \xmark & \xmark & \xmark & \cmark & \xmark & \xmark & \cmark & \cmark & \cmark & –  \\ 
Han et al.\,\cite{Han2024KANSY}           & \xmark & \xmark & \xmark & \xmark & \xmark & \xmark & \xmark & \xmark & \xmark & \xmark & 3  \\ 
Mohan et al.\,\cite{mohan2024kans}          & \cmark & \xmark & \xmark & \xmark & \cmark & \xmark & \xmark & \cmark & \cmark & \cmark & 5  \\ 
Yu et al.\,\cite{Yu2024ExploringKN}                 & \xmark & \xmark & \xmark & \xmark & \xmark & \xmark & \xmark & \xmark & \xmark & \xmark & 6  \\ 
Cang et al.\,\cite{Cang2024CanKW}            & \cmark & \xmark & \xmark & \xmark & \xmark & \xmark & \xmark & \cmark & \cmark & \cmark & 5  \\ 
Abd Elaziz et al.\,\cite{abdaizaz}    & \cmark & \xmark & \xmark & \xmark & \xmark & \xmark & \xmark & \cmark & \cmark & \cmark & 2  \\ 
Livieris et al.\,\cite{math12193022}    & \cmark & \xmark & \xmark & \xmark & \xmark & \xmark & \xmark & \xmark & \xmark & \xmark & –  \\ 
Cheon et al.\,\cite{cheon2024kolmogorov}       & \xmark & \xmark & \xmark & \xmark & \xmark & \xmark & \xmark & \cmark & \cmark & \cmark & 6  \\ 
Ferdaus et al.\,\cite{Ferdaus2024KANICEKN}      & \cmark & \xmark & \xmark & \xmark & \cmark & \xmark & \xmark & \cmark & \cmark & \cmark & 2  \\ 
\hline
\textbf{Ours}                                & \cmark & \cmark & \cmark & \cmark & \cmark & \cmark & \cmark & \cmark & \cmark & \cmark & \textbf{24} \\
\hline
\end{tabular}
\begin{tablenotes}\footnotesize
\item “Time” denotes either training time or per-sample inference latency as reported.  “Sweeps” counts distinct hyper-parameter configurations in the authors’ ablation study.
\end{tablenotes}
\end{threeparttable}
\end{table*}

\begin{equation}
K_{i,j} = \Phi_{i,j} \text{ where } \Phi_{i,j} \text{ is a matrix of B-splines of size } n_2, n_1
\label{eq:kernel_def}
\end{equation}

Then from the definition of a CNN \cite{lecun1989backprop}, we can write the $(i,j)^{th}$ entry of the feature map is given by the following general formula:

\begin{equation}
(image * k)_{i,j} = \sum_{x}^{m_1}\sum_{y}^{m_2}\Phi_{xy}(a_{i-x,i-y})
\label{eq:feature_map}
\end{equation}

The authors indicate generalization of their hypothesis but don't offer complete generalization. Although by evaluation of their starting hypothesis, we deduce it to be the following:

\begin{equation}
K = \begin{bmatrix} 
\Phi_{11} & \Phi_{12} & \cdots & \Phi_{1n_1} \\
\Phi_{21} & \Phi_{22} & \cdots & \Phi_{2n_1} \\
\vdots & \vdots & \ddots & \vdots \\
\Phi_{n_21} & \Phi_{n_22} & \cdots & \Phi_{n_2n_1}
\end{bmatrix}
\label{eq:k_matrix}
\end{equation}

\begin{equation}
I = \begin{bmatrix}
a_{11} & a_{12} & \cdots & a_{1m_2} \\
a_{21} & a_{22} & \cdots & a_{2m_2} \\
\vdots & \vdots & \ddots & \vdots \\
a_{m_11} & a_{m_12} & \cdots & a_{m_1m_2}
\end{bmatrix}
\label{eq:image_matrix}
\end{equation}

Then, $I \times K$ (Image $\times$ Kernel) would be given by the following:

\begin{equation}
\small
I \times K = \begin{bmatrix}
\sum\limits_{i=1}^{n_2}\sum\limits_{j=1}^{n_1} \!\Phi_{ij}(a_{1,j+(i-1)n_1}) & \!\cdots\! & r_1(m_2\!-\!1) \\[2ex]
\sum\limits_{i=1}^{n_2}\sum\limits_{j=1}^{n_1} \!\Phi_{ij}(a_{2,j+(i-1)n_1}) & \!\cdots\! & r_2(m_2\!-\!1) \\[2ex]
\vdots & \ddots & \vdots \\[1.5ex]
\sum\limits_{i=1}^{n_2}\sum\limits_{j=1}^{n_1} \!\Phi_{ij}(a_{m_1,j+(i-1)n_1}) & \!\cdots\! & r_{m_1}(m_2\!-\!1)
\end{bmatrix}
\label{eq:convolution}
\end{equation}

Where $r_{m_1}(m_2\text{-}1)$ denotes continuation of operation until the end of the row for $m_2-1$ columns.

The authors from \cite{bodner2024convolutional} experimented with their proposed model on MNIST and Fashion MNIST \cite{lecun1998mnist, xiao2017fashion}. The results from authors of \cite{bodner2024convolutional} contradicts the results proposed by the authors at \cite{yu2024kan}. The authors \cite{bodner2024convolutional} concluded that a Convolutional KAN outperformed a traditional CNN with a KAN based kernel with similar number of parameters in terms of accuracy, precision, recall and F1 score.

This contradiction in results between the two authors could be expected because CKAN brings new concepts in computer vision. Therefore, the field of Convolutional KANs calls for more research with more complex and higher standard datasets. 

\begin{figure*}[t]
\centering
\includegraphics[width=\linewidth]{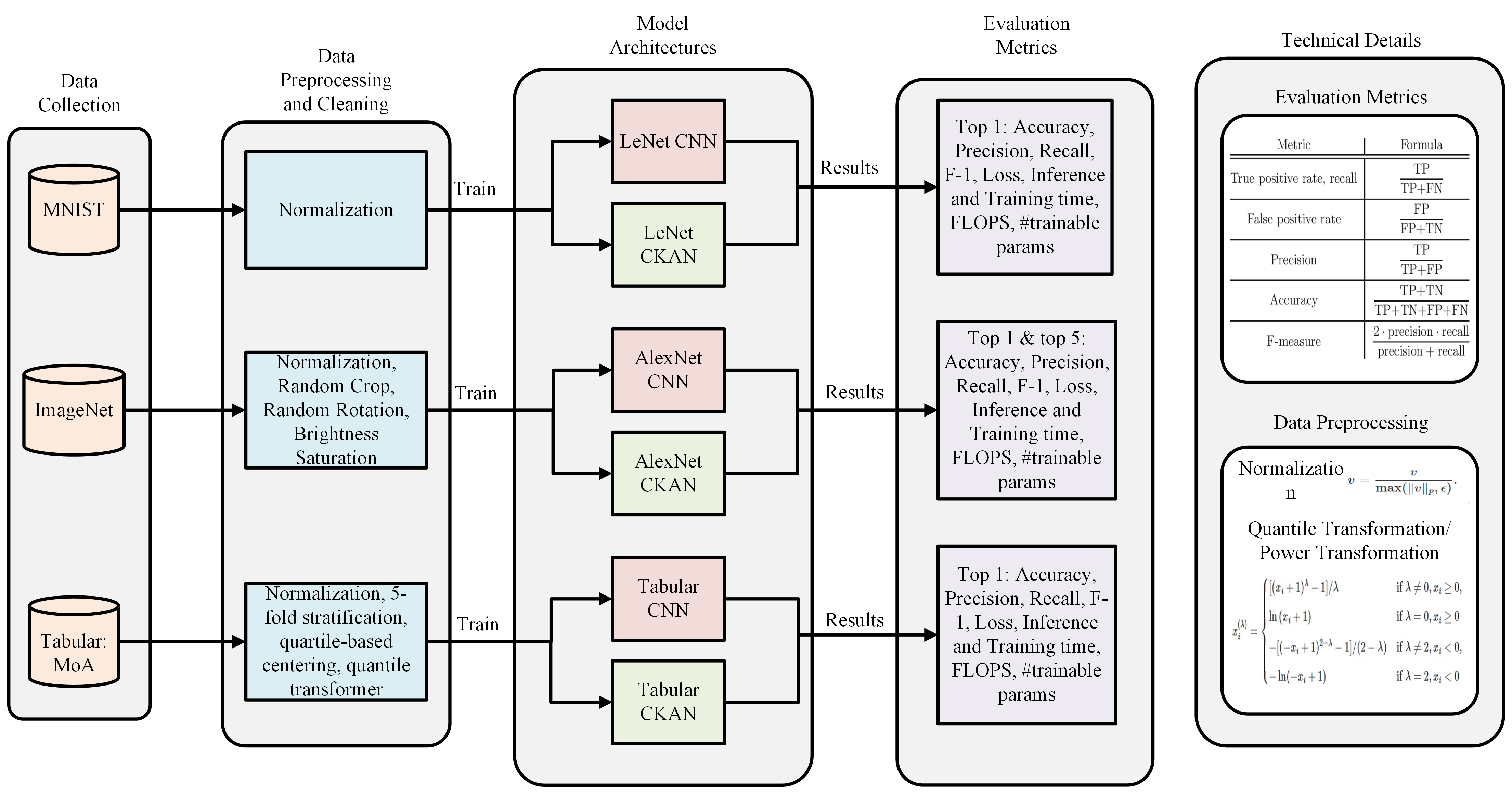}
\caption{Summary of Research Methodology}
\label{fig:methodology}
\end{figure*}

A condensed summary of current research on CKAN is highlighted in table \ref{tab:research_summary}. We note a need for a paper to address not only the future promise of KAN on small dataset but also test it's feasibility on larger research projects across computer vision and tabular settings; this paper fulfills the research gap presented by providing an in depth comparison of CKANs in the three datasets that highlight the three needs of modern research: fast prototype (easy to build), interoperability of results across bigger and diverse dataset (easy to scale) and fast inference (easy to deploy). 

\section{Methodology}\label{Section:methodology}
Our research procedure is highlighted on the Fig. \ref{fig:methodology}. Based on the Fig. \ref{fig:methodology} and the main objective of the paper, the methodology section is best divided into three parts: Computer Vision, Tabular Classification and Evaluation Metrics, each equipped with their own subsection discussing data preprocessing and hyperparameter selection.

\subsection{Computer Vision}
Two architectures based on popular CNN architectures, namely LeNet and AlexNet, were trained using KAN based convolution layers; for convention let's name them LeNet KAN and AlexNet KAN. The only difference between the AlexNet and LeNet architecture from their CKAN counterparts is the number of filters while the overall architecture remains same.


\subsubsection{Dataset and preprocessing for Computer Vision}\label{sec:compvis_data}
To maintain consistency with the original paper LeNet KAN was trained on the MNIST~\cite{lecun1998mnist} and AlexNet KAN was trained on the ImageNet~\cite{deng2009imagenet} dataset. The MNIST dataset is a collection of 60,000 grayscale handwritten digits belonging to 10 classes whereas the ImageNet is the collection of 1.2 million images belonging to 1000 classes. In order to maintain the integrity of the original research and a fair comparison between the KAN and MLP counterpart, we didn't do additional preprocessing on either of the dataset rather than to normalize them, which was automatically handled by pytorch.

\subsubsection{Hyperparameter Selection}\label{subsec:hyperpara_compvis}
For a standard CKAN by~\cite{bodner2024convolutional} the number of parameters as compared to that of a standard CNN by equations~\ref{eq:kan_params} and~\ref{eq:mlp_params} would be four times higher. Since the LeNet architecture doesn't have all number of filters in the order of $2^n$, the first layer had to be rounded up to maintain at least 2 filters on the layer. Adam~\cite{kingma2014adam} was chosen as the optimizer with the same learning rate.

Similarly, reducing the number of filters by one fourth would lead to the AlexNet KAN to have extremely less number of filters. To give the best of advantages to the hypothesis of authors~\cite{bodner2024convolutional, liu2024kan} we adjust the number of filters by the best effort to keep the number of trainable parameters within the same range. Although it is contended that CKAN can perform better than CNN with fewer number of parameters, we gave the best chance to both of the models and present their result alongside the number of parameters the models had. AlexNet KAN was let to train for 100 epochs with early stopping on tolerance 3 for the validation loss whereas the LeNet models were let to train for 50 epochs with early stopping on tolerance 3 for the same. We use PyTorch's pretrained AlexNet for CNN's part as it is the industry standard.

\subsection{Tabular CNN}
Guo~\cite{guo2020kaggle} has shown their 1-Dimensional CNN architecture to be used on a tabular dataset, which also won the second prize on the Kaggle's MOA competition ~\cite{paik2020moa}. Given that tabular dataset doesn't have similar properties next to one another and CNNs are built to recognize similar patterns next to one another, we follow their specific implementation where the input table row is projected into a vector where each of the adjacent data comes as a closely related feature representation of one entry in the table row. Then we apply the 1-D CNN layer on the generated vector which would make sure that no adjacent data of vastly different nature would have to go through the CNN filter at once as a closely related pattern.

\begin{figure*}[t]
\centering
\includegraphics[width=\linewidth]{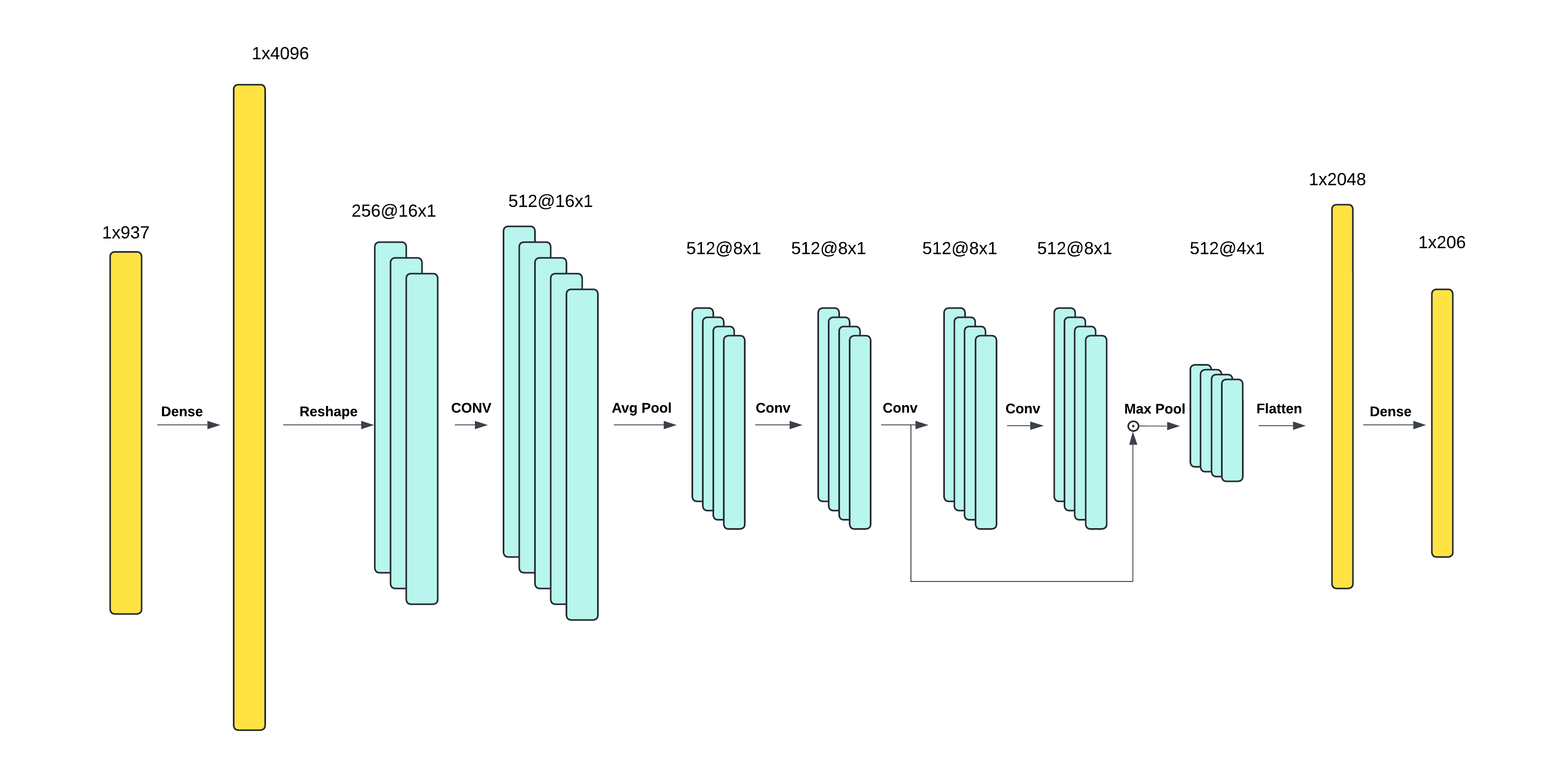}
\caption{1D CNN architecture}
\label{fig:1d_cnn}
\end{figure*}

\begin{figure*}[t]
\centering
\includegraphics[width=\linewidth]{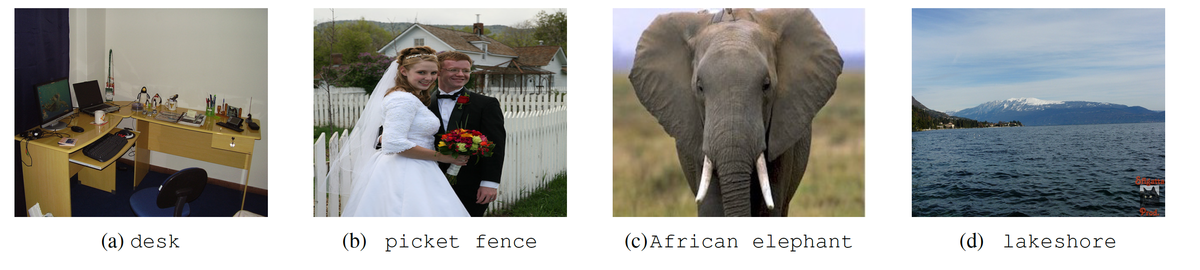}
\caption{Imagenet Sample from~\cite{waterloo2024categories}}
\label{fig:imagenet_sample}
\end{figure*}

\subsubsection{Dataset for Tabular Classification}
In order to replicate the same result as Guo~\cite{guo2020kaggle}, the same dataset Mechanisms of Action (MoA)~\cite{paik2020moa} was chosen for this task. In order to have a fairer comparison, all the data preprocessing steps followed by Guo were also replicated for the KAN based architecture.

\subsubsection{Hyperparameter Selection}
Since reducing the number of filters by one fourth in this case would not have a significant impact on the model's width or depth, we decided to match the number of parameters and reduce the number of layers in each layer in KAN by one fourth. This would result in the model architecture for Tabular CKAN to be the same as Fig.~\ref{fig:1d_cnn}, just with fewer (one fourth) number of filters or kernels in it's convolutions.

The approach for the loss function chosen was also with the same rationale. Custom weighted loss function proposed by Guo~\cite{guo2020kaggle} was used on both the models with the same learning rate and weight decay for the Adam~\cite{kingma2014adam} optimizer.

These hyperparameters are more aligned towards direct comparison of a CKAN vs CNN in a single dimension, whereas the hyperparameters chosen in the section \ref{subsec:hyperpara_compvis} may tend to favor a alight higher number of trainable parameters for CKAN over CNN (in LeNet) but this decision wouldn't ultimately matter because the models fail to justify the higher learning capacity.

\begin{figure*}[t]
    \centering
\includegraphics[width=\linewidth]{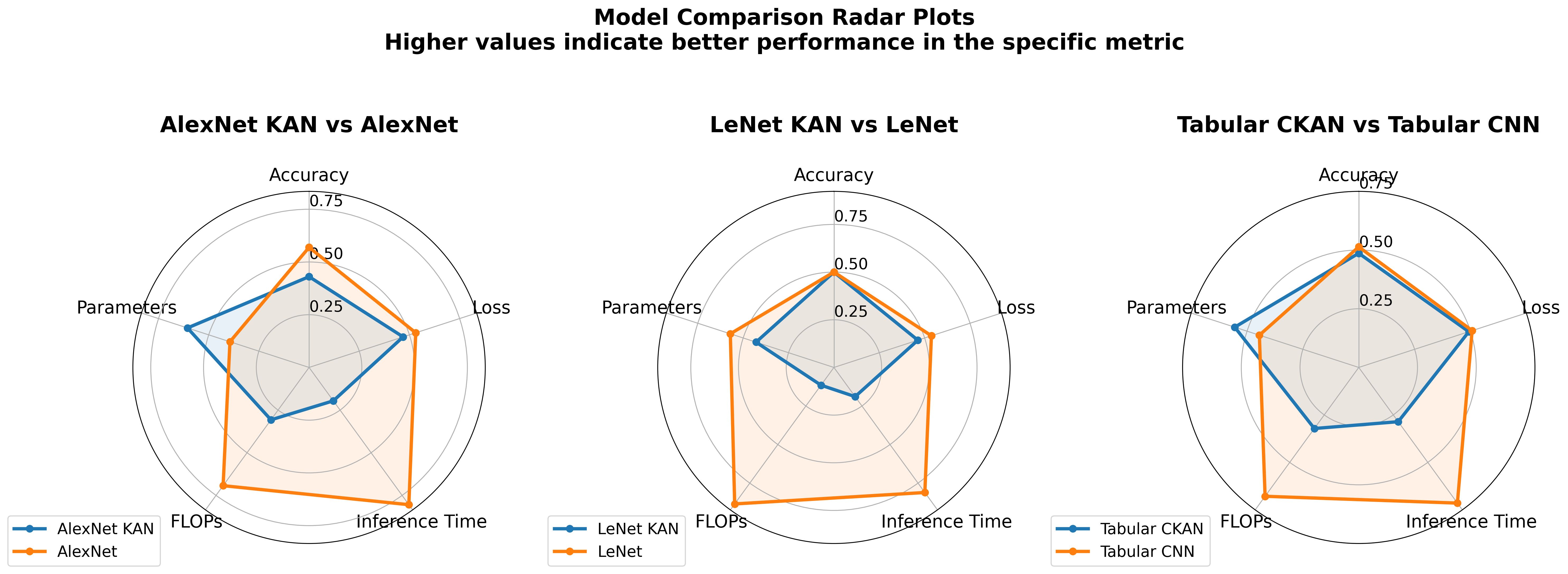}
\caption{Radar Plot of Performance Metrics of All Tested Models. All metrics except accuracy are inverted after normalization.}
\label{fig:radar}
\end{figure*}

\subsection{Evaluation Metrics}
Given the constraint of 1000 classes in the ImageNet dataset, it is simply not possible to treat it as the same as other relatively smaller dataset. Therefore there is a need to find metrics beyond classification matrix and report that encapsulates the depth and complexities of the results produced by the models. The following metrics were deployed to evaluate the results produced by the models.

\subsubsection{Top-5 Accuracy, Precision, Recall and F-1 Score}
We treat the top 5 classes in the output probability distribution as the predicted classes, i.e., if any one of the classes in the top 5 probability distribution is true then the result is considered true. Accuracy is then calculated similarly as the percentage of correct predictions in the dataset. Such a metric becomes necessary in ImageNet because for any given image in the dataset, it could have multiple outputs which could all be plausible. Consider Fig.~\ref{fig:imagenet_sample} for example, Fig.~\ref{fig:imagenet_sample}a could have been labeled as a keyboard, chair, computer or pens and all would have been correct. Similarly for Fig.~\ref{fig:imagenet_sample}b, the prediction could have been a bride, groom, wedding dress, or flowers and all would have been correct. Thus, it is deemed that considering the Top-5 Accuracy would serve as a necessary while evaluating such a complex dataset.

\begin{equation}
Top-1/5-Accuracy = \frac{TP + TN}{TP + TN + FP + FN}
\label{eq:accuracy}
\end{equation}
\begin{equation}
Top-1/5-Precision = \frac{TP}{TP + FP}
\label{eq:precision}
\end{equation}

\begin{equation}
Top-1/5-Recall = \frac{TP}{TP + FN}
\label{eq:recall}
\end{equation}

\begin{equation}
Top-1/5-F1\text{-}Score = \frac{2 \cdot Precision \cdot Recall}{Precision + Recall}
\label{eq:f1}
\end{equation}

Where TP = True Positives, FP = False Positives, TN = True Negatives and FN = False Negatives.

\subsubsection{Top-1 Accuracy, Precision, Recall and F-1 Score}
Also commonly just termed as accuracy, Top-1 accuracy treats the class with the highest probability in the output layer's probability distribution as the predicted class and then calculates the percentage of correct predictions in the dataset. Similar to Top-1 Accuracy, we would treat the class with the highest probability in the final output layer as the predicted class and then calculate the Precision, Recall and F1-Score of the model based on eq \ref{eq:accuracy}-\ref{eq:f1}.

\subsubsection{Cross-Entropy Loss}
The Cross-Entropy loss function measures the difference between discovered probability distribution of a classification model and the predicted values. This is used for non-binary outcomes where we sum over M classes for N times.

\begin{equation}
Cross\text{-}Entropy = -\sum_{i=1}^{N}\sum_{j=1}^{M}y_{i,j}\log(p(y_{i,j}))
\label{eq:cross_entropy}
\end{equation}

Where $y_{i,j}$ is the actual label of a data and $p(y_{i,j})$ is the model's probability distribution for the class.

\subsubsection{Binary Cross-Entropy (BCE) Loss}
The authors of MoA dataset~\cite{paik2020moa} chose to do BCE loss over M classes instead of Cross-Entropy loss. In order to maintain consistency and get results for the hidden test cases, we would use the same metric as the authors of~\cite{paik2020moa} had set for the authors of~\cite{guo2020kaggle} in order to do tabular classification using convolutional models.

\begin{equation}
\begin{split}
BCE = -\frac{1}{M}\sum_{i=1}^{M}\frac{1}{N}\sum_{j=1}^{N}[&y_{i,j}\log(p(y_{i,j})) \\ 
    &+ (1-y_{i,j})\log(1-p(y_{i,j}))]
\end{split}
\label{eq:bce}
\end{equation}

\setlength{\tabcolsep}{6pt}
\renewcommand\arraystretch{1.15}

\begin{table}[h!]
  \centering
  \caption{Design matrix for the 24-run ablation sweep.
           \textcolor{RoyalBlue}{$\blacktriangle$} = default level.
           Parameter counts are theoretical pre-prune values.}
  \label{tab:fancy_ablate_grid}

  \begin{tabular}{@{}l c c c c@{}}
    \toprule
    \multirow{2}{*}{\textbf{Factor}} &
    \multicolumn{3}{c}{\textbf{Levels}} &
    \multirow{2}{*}{\makecell{\textbf{Params}\\[-2pt](X10\textsuperscript{3})}} \\ \cmidrule(lr){2-4}
    & \textbf{L1} & \textbf{L2} & \textbf{L3} & \\ \midrule
    Spline grid $g$ &
      4\,\textcolor{RoyalBlue}{$\blacktriangle$} &
      8 &
      16 &
      72–179 \\[2pt]

    Width mult.\ $w$ &
      1.0\,\textcolor{RoyalBlue}{$\blacktriangle$} &
      1.5 & – & 72–179 \\[2pt]

    ReLU &
      On\,\textcolor{RoyalBlue}{$\blacktriangle$} &
      Off & – & — \\[2pt]

    Prune ratio $p$ &
      0\,\% &
      25\,\%\,\textcolor{RoyalBlue}{$\blacktriangle$} & – &
      69–155 \\

    \bottomrule
  \end{tabular}
  \vspace{-6pt}
\end{table}

\subsection{Ablation-Study Protocol}\label{sec:ablation}
The ablation study protocol grid is presented in \cref{tab:fancy_ablate_grid}. In order to make the study viable in terms of compute hours, we choose to do the 24-run ablation study in the LeNet architecture. We specifically avoid doing it on AlexNet architecture because (more discussed on \cref{section:results}), a 24 sweep training run in the ImageNet data would cost us approximately 1150 GPU days in CKAN which is a significant investment. Three key ideas not summarized by the \cref{tab:fancy_ablate_grid} are shortly discussed:

\paragraph{Dataset \& Training Recipe} We used the same set of MNIST dataset (60k training/ 10k testing/validation) as previously discussed in \cref{sec:compvis_data} with identical normalization. The experiments were done with Adam ($\eta\!=\!10^{-3}$), batch 512, 5 epochs and a single NVIDIA P100. This ensured all the 24 runs were completed within 1 hour while establishing a strong foundation to draw ablation conclusions.

\paragraph{Metrics} We log validation loss, accuracy, parameter count, multiply–accumulate operations (MACs, via THOP), and batch-32 latency measured with CUDA events. This provides a well detailed analysis of the efficiency factors including the tradeoffs between efficiency and accuracy of the architecture.

\paragraph{Pruning Implementation} Channel-wise $L_{2}$ structured pruning \cite{hanCKAN} is applied to every internal Conv2d in FastKANConv2DLayer.
We keep a 25\,\% sparsity mask fixed during fine-tuning. This strategy on ablation would determine the generalizability of the results obtained from \cref{Section:methodology}.

\subsection{Hyperparameters and Training Information}
The models were then trained on the University of Southern Mississippi's High Performance Cluster (HPC) Magnolia with the Nvidia Tesla K80 GPU and Nvidia Tesla P100 GPU. A total of 4 $\times$ K80 GPUs were utilized to train the AlexNet KAN model whereas for the LeNet counterparts 2 $\times$ P100 GPUs were utilized. Training for both the computer vision models required the utilization of Distributed Data Parallel (DDP) computing using pytorch in order to increase the batch size to a considerable amount; in our case 16 for the ImageNet dataset. For the tabular data, we trained both models on a single kaggle notebook with accelerated 1 $\times$ P100 GPU.

\section{Results} \label{section:results}

The results from the metrics evaluated as per the methodology fig \ref{fig:methodology} is presented on this section of the paper. We note that before evaluating the results of each models, we need to evaluate the inference and training efficiency of the approaches. Thus, the section is presented on the following way: section \ref{efficiency} discusses the efficiency of the models including FLOPS, training and inference time, and number of trainable parameters, section \ref{tabular}, \ref{alexnet}, \ref{lenet} discusses the rest of the metrics like accuracy, precision, recall and f-1 score for Tabular CNN, AlexNet and LeNet similarly.  

\begin{table*}[!ht]
\caption{Evaluation metrics of AlexNet KAN, AlexNet, LeNet KAN, LeNet, Tabular CNN, Tabular CKAN}
\label{tab:evaluation_metrics}
\centering
\renewcommand{\arraystretch}{1.2}  
\begin{tabular}{l*{6}{c}}  
\toprule
\textbf{Evaluation Metrics} & \makecell{\textbf{AlexNet} \textbf{KAN}} & \textbf{AlexNet} & \makecell{\textbf{LeNet} \textbf{KAN}} & \textbf{LeNet} & \makecell{\textbf{Tabular} \textbf{CNN}} & \makecell{\textbf{Tabular} \textbf{CKAN}} \\
\midrule
Top-5 Accuracy    &67.72& 79.07& -- & -- & -- & -- \\
Top-5 Precision   &67.92 & 78.66& -- & -- & -- & -- \\
Top-5 Recall      &67.72 & 79.07& -- & -- & -- & -- \\
Top-5 F1 Score    &66.02 & 78.00& -- & -- & -- & -- \\
\midrule
Top-1 Accuracy    &42.79 & 56.62 &98.81 & 98.89&47.61 &45.09 \\
Top-1 Precision   &42.91 &56.29 &98.79 &  98.88&94.66 & 98.13\\
Top-1 Recall      &42.79 &56.62 &98.79 &  98.87&26.95 &24.30\\
Top-1 F1 Score    &41.72 &55.75 & 98.79 & 98.88&28.36 &25.29 \\
\midrule
Loss (BCE \& CE)  &2.62 &2.31 & 0.036 & 0.031 & 0.0167 & 0.0172 \\
Inference Time    &0.0074s &0.0018s & 0.003s & 0.0007s &0.00004s & 0.0001s \\
FLOPS             &1,611,568,352 &714,197,696 &3,298,728 &429,128 &37,853,010 &798,61,586 \\
Training Time     &48 days &3 days& 981.45s &888.77s  &6450.5s &10646.07s \\
\# of Parameters  &39,756,776 &61,100,840 &82,128 &61,750 & 7818482 & 6265998\\
\bottomrule
\end{tabular}
\end{table*}

\subsection{Inference and Training Time}\label{efficiency}
The first and most important consideration in this section is time efficiency. The entire theme of the results observed during the experiment revolved around the throughput of CKAN compared to CNN per unit time. Within the context of the paper, we gathered all metrics presented in Table \ref{tab:evaluation_metrics} and normalized the metrics to compare and contrast on how each of them performed against each another and present on fig \ref{fig:radar}. It is evident that the discussion to be followed is regarding the poor performance of the KAN models. For the plot Loss, FLOPS, Inference Time and Parameters are inverted after normalization to show the widespread dominance of CNN models against CKAN. Even while having lesser number of parameters in AlexNet KAN and Tabular CKAN, the inference time in the radar plot shows the models are not optimized enough for modern classification tasks fig \ref{fig:radar}.

The AlexNet KAN model took 48 days to train for 100 epochs. In contrast, the standard AlexNet model, by hand calculation of the original report, would have required at max (in worst case scenario) 3 days time to complete the same number of epochs. This represents an increase of almost 16 times in time consumption for the KAN model. While the inference time for single image is 4 times more in the KAN counterpart of the same architecture (0.0074s vs 0.0018s), the number of trainable parameters is just 63\%. This indicates that even with less number of parameters the inference is terrible on AlexNet KAN. Even so for the FLOPS, where it is more than 2x that of AlexNet on PyTorch (1,611,568,352 vs 714,197,696).

Similarly, for the LeNet counterpart, the LeNet KAN model took 981.45s to complete 50 epochs, while the standard LeNet model took 888.77s for the same number of epochs. This is a time difference of 92.68s. The inference time for a single input was calculated to be 0.003s and 0.0007s for the LeNet KAN and LeNet respectively. This shows an increase of ~3.28x just on single image inference time. Similar is the case for the FLOPS count; with just a few thousands more number of trainable parameters from the b splines, 82,128 in LeNet KAN as compared to 61,750 in LeNet, the FLOPS increased by 7 times from 429128 in LeNet to 3298728 in LeNet KAN.

A comparable pattern was observed for the simpler Tabular CNN models. The lightweight Tabular CNN KAN model required 10646.07s to train for 15 seeds, 4 folds of 24 epochs each, while the standard lightweight Tabular CNN model only needed 6450.5s for the same setting. This shows that the KAN model took almost 1.65 times longer. As with the other models the inference time was also lagging behind with the Tabular CNN finishing single inference at 0.00004s and the CKAN counterpart at 0.0001s. Interesting observation to be noted from the Table \ref{tab:evaluation_metrics} is even with ~1.25 times more number of trainable parameters, the Tabular CNN performed better in FLOPS (2x less), inference and overall training time.

The rest of the parameters in Table~\ref{tab:evaluation_metrics} are discussed per model in the following sections.

\subsection{Tabular~CNN versus Tabular~CKAN}\label{tabular}
The Tabular experiment was carried out on a relatively small, structured
biological dataset whose statistical complexity is modest compared with
natural images.  
Both architectures achieved high scores, yet the
baseline Tabular‐CNN remained ahead in the core classification metrics:
accuracy, recall and F\textsubscript{1} were
47.61\,\%, 26.95\,\% and 28.36\,\%, respectively, whereas the
CKAN variant reached 45.09\,\%, 24.30\,\% and 25.29\,\%.
The only metric favoring CKAN was precision
(98.13\,\% versus 94.66\,
strong regularizer in the positive class but do not translate that
advantage into overall recognition performance.  
Given that the CKAN model also incurred a 1.6\,X longer training time and a
2.5\,X slower inference rate (Section~\ref{efficiency}),
the Tabular‐CNN provides a more favorable accuracy–efficiency trade-off
for this type of data.

\subsection{LeNet‐KAN versus LeNet}\label{lenet}
On the \textsc{MNIST} benchmark the KAN adaptation of LeNet performs
comparably to its convolutional counterpart, reflecting the relatively small
input resolution and class set.  
Accuracy, precision, recall and F\textsubscript{1} differ by less than
0.1\,pp across the two models
(98.81, 98.79, 98.79, 97.79 for LeNet‐KAN
versus 98.89, 98.88, 98.87, 98.88 for LeNet).
Despite this metric parity, the KAN variant is markedly less efficient:
training time rose from 889 s to 981 s and single-image inference latency
from 0.7\,ms to 3.0\,ms, while FLOPs increased seven-fold.  
The result underscores a recurring pattern: CKANs can match CNN accuracy on
simple images, but presently do so only at a disproportionate computational
cost.

\subsection{AlexNet‐KAN versus AlexNet}\label{alexnet}
The gap widens on the far more demanding ImageNet-1k task.  
Across Top-1 and Top-5 metrics the AlexNet‐KAN trails standard AlexNet by
14–16\,pp, despite exhibiting roughly double the FLOP count and 
four times the inference latency documented in
Table~\ref{tab:evaluation_metrics}.
Specifically, the KAN model attains 42.8\,\% / 67.8\,\%
Top-1 / Top-5 accuracy, whereas the convolutional baseline reaches
56.6\,\% / 79.1\,\%.
These results suggest that the spline–based representation does not
capture the hierarchical abstractions required for large-scale vision
and that further architectural or optimization advances are necessary
before CKANs can serve as a practical replacement for deep CNNs in this
regime.

\begin{figure*}[!t]   
  \centering
  \begin{subfigure}[b]{0.32\textwidth}
    \centering
    \includegraphics[width=\linewidth]{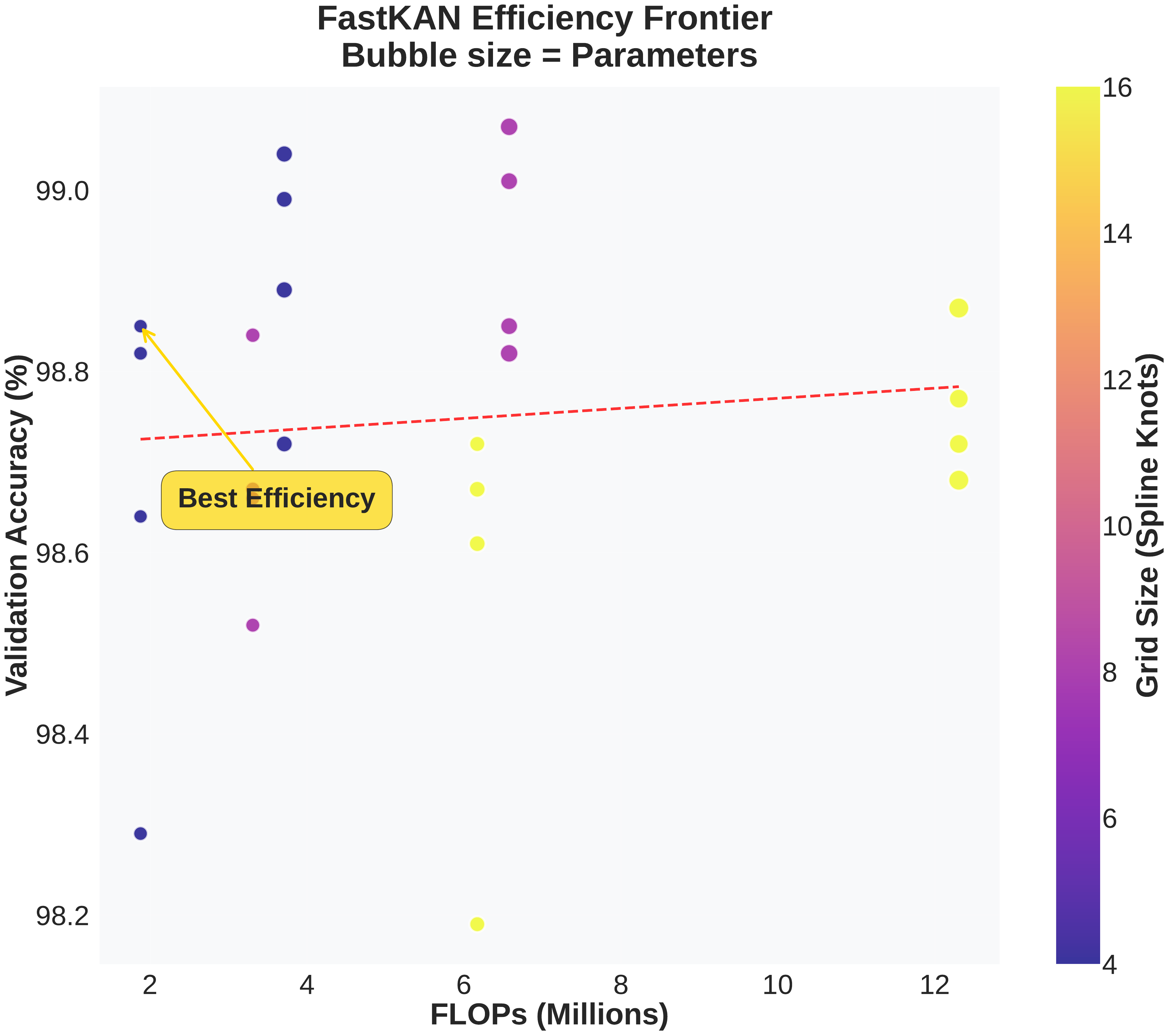}
    \caption{Acc. vs FLOPS efficiency.}
    \label{fig:ablate_acc_grid}
  \end{subfigure}\hfill
  \begin{subfigure}[b]{0.32\textwidth}
    \centering
    \includegraphics[width=\linewidth]{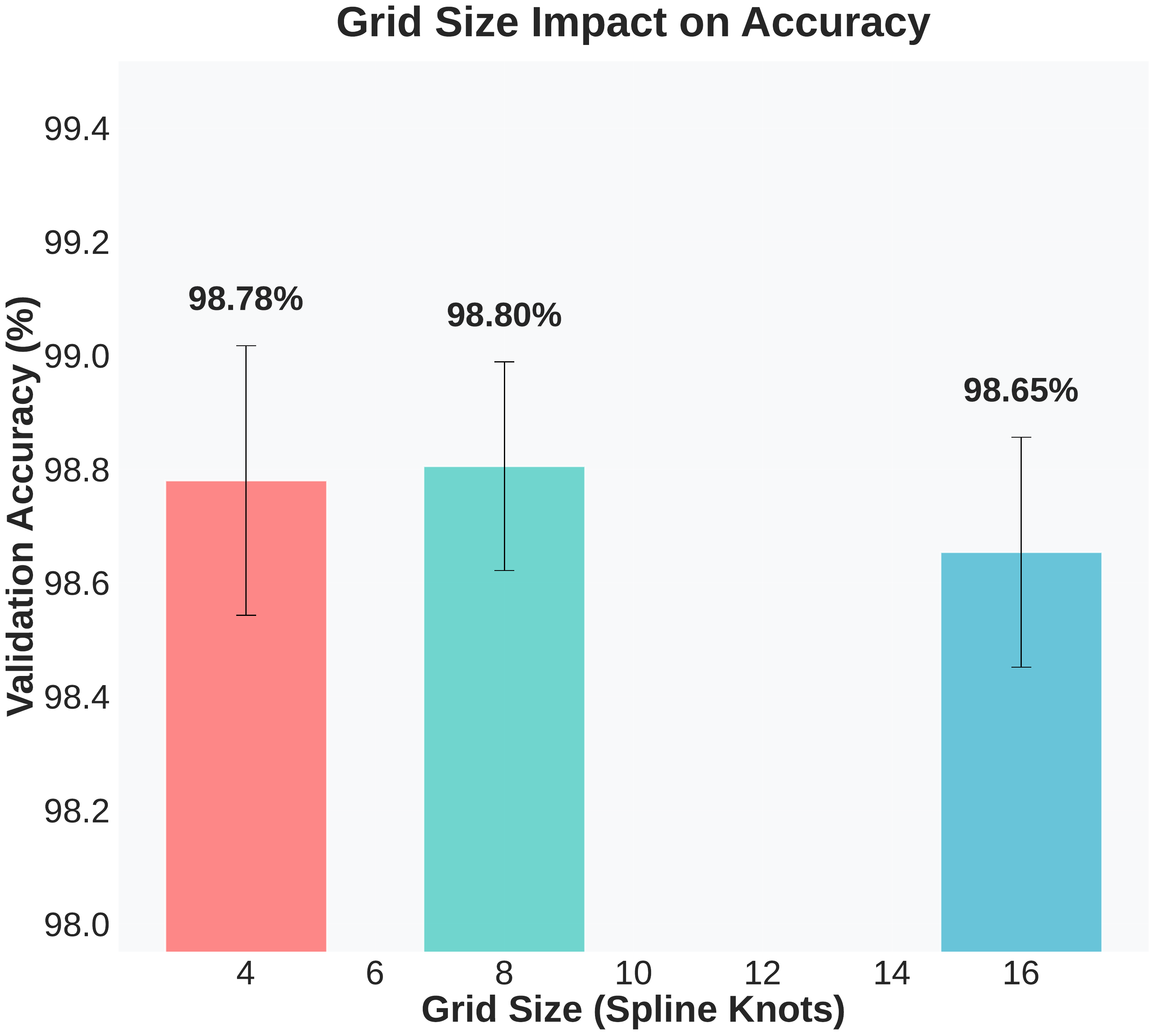}
    \caption{Grid size vs Validation Acc.}
    \label{fig:ablate_lat_flops_fp32}
  \end{subfigure}\hfill
  \begin{subfigure}[b]{0.32\textwidth}
    \centering
    \includegraphics[width=\linewidth]{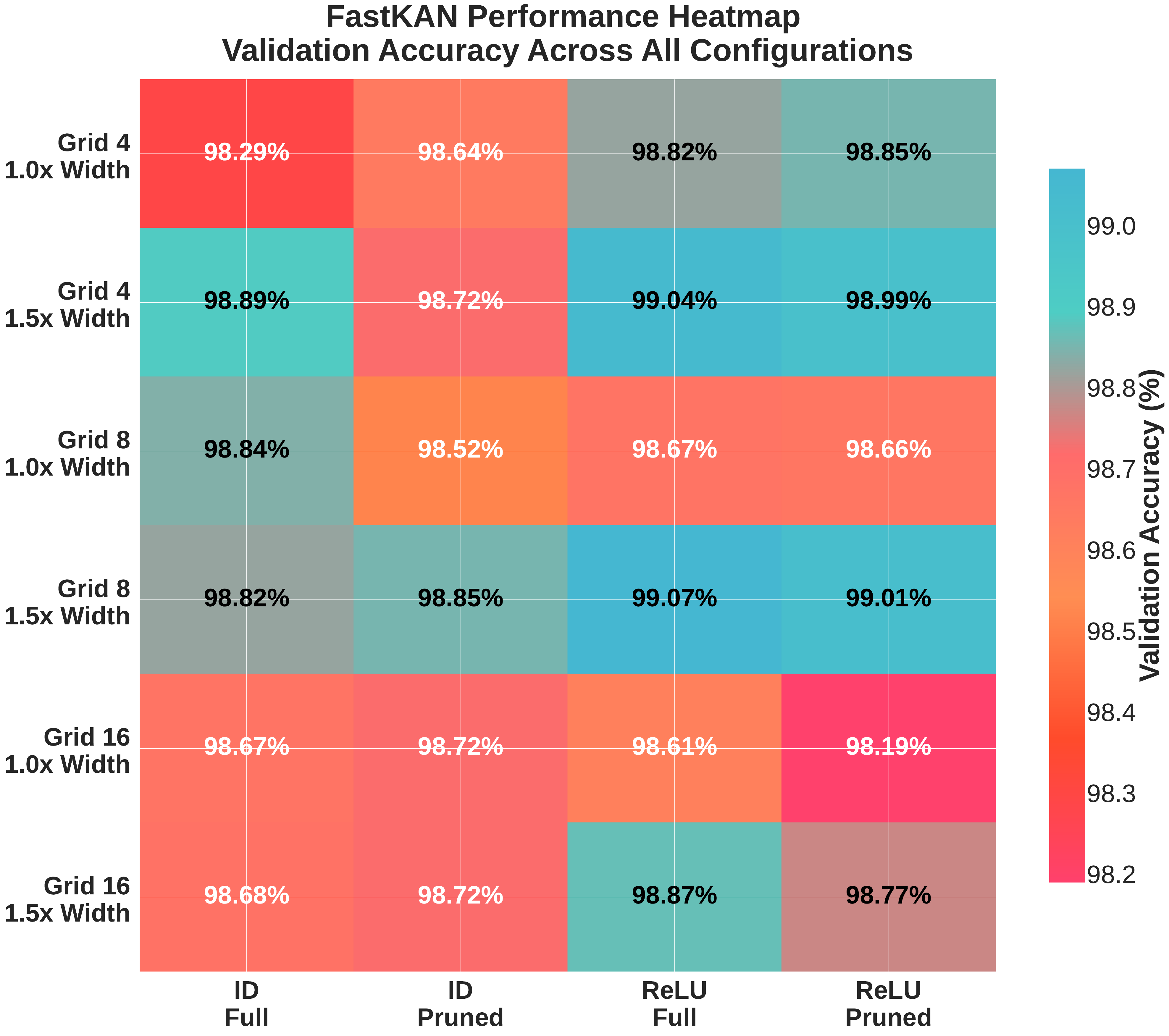}
    \caption{Validation Accuracy heatmap against all conditions.}
    \label{fig:ablate_lat_flops_pruned}
  \end{subfigure}

  \vspace{0.7em} 

  \begin{subfigure}[b]{0.32\textwidth}
    \centering
    \includegraphics[width=\linewidth]{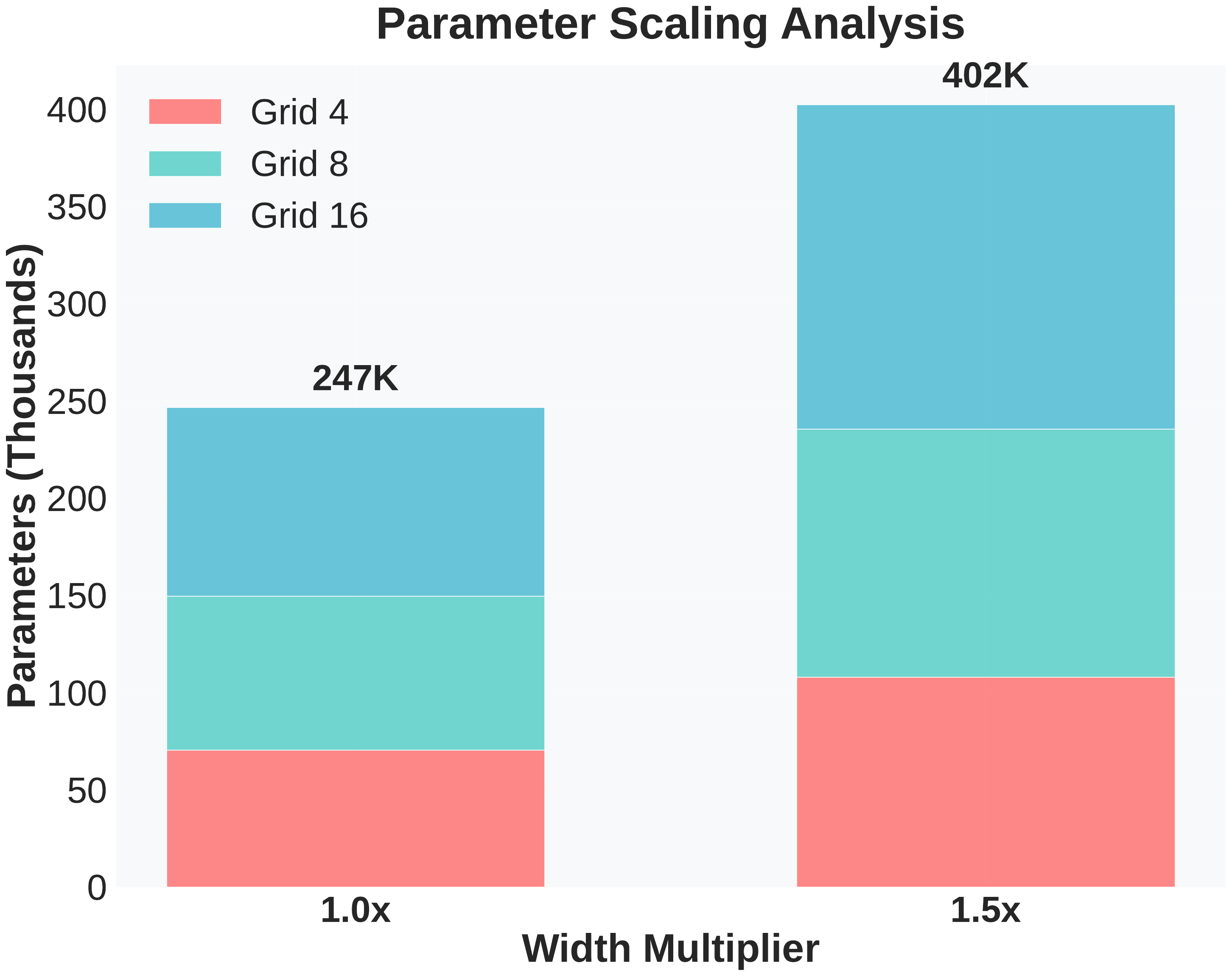}
    \caption{Parameter scaling with width multiplier.}
    \label{fig:ablate_acc_lat}
  \end{subfigure}\hfill
  \begin{subfigure}[b]{0.32\textwidth}
    \centering
    \includegraphics[width=\linewidth]{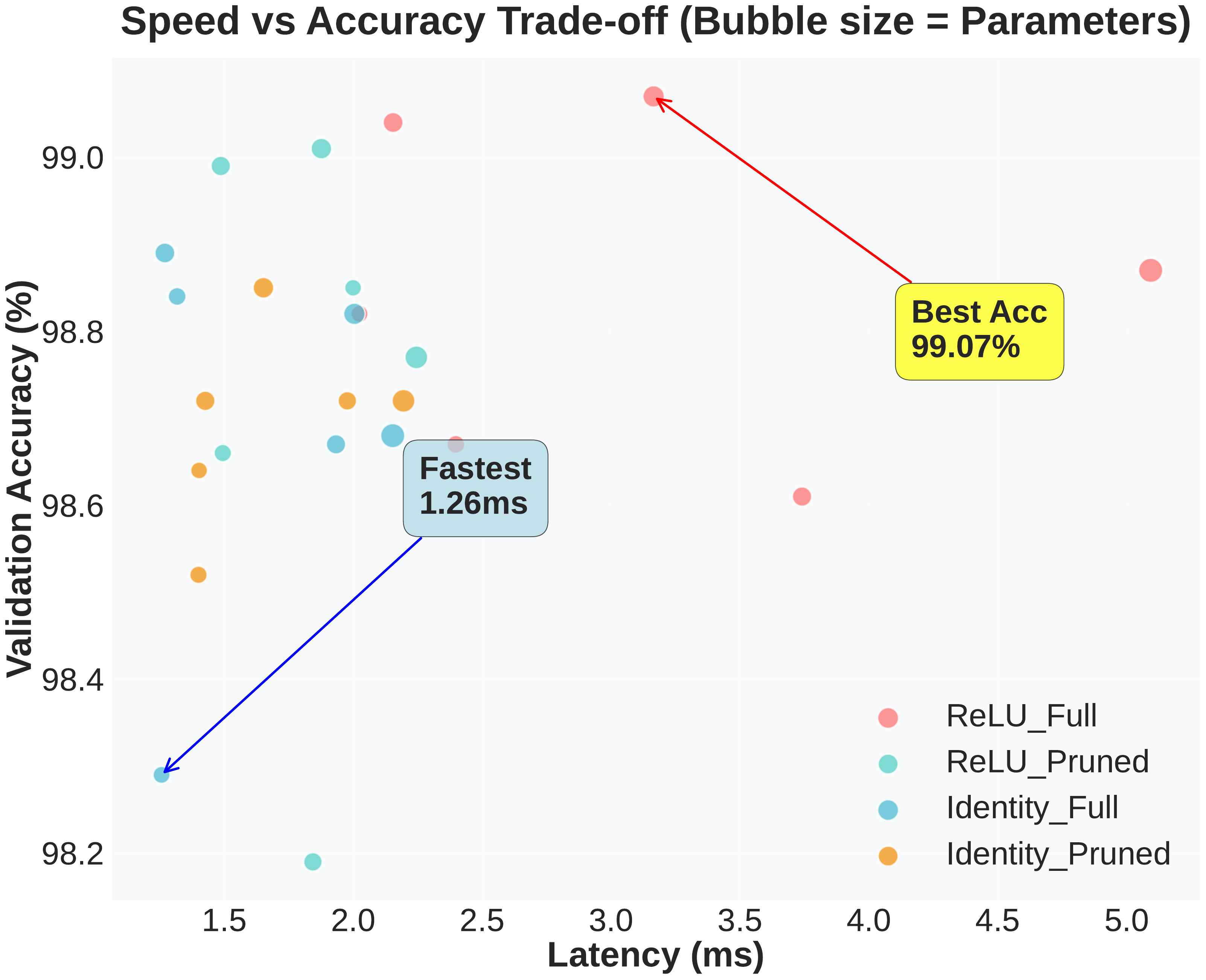}
    \caption{Latency vs performance analysis.}
    \label{fig:ablate_params_acc}
  \end{subfigure}\hfill
  \begin{subfigure}[b]{0.32\textwidth}
    \centering
    \includegraphics[width=\linewidth]{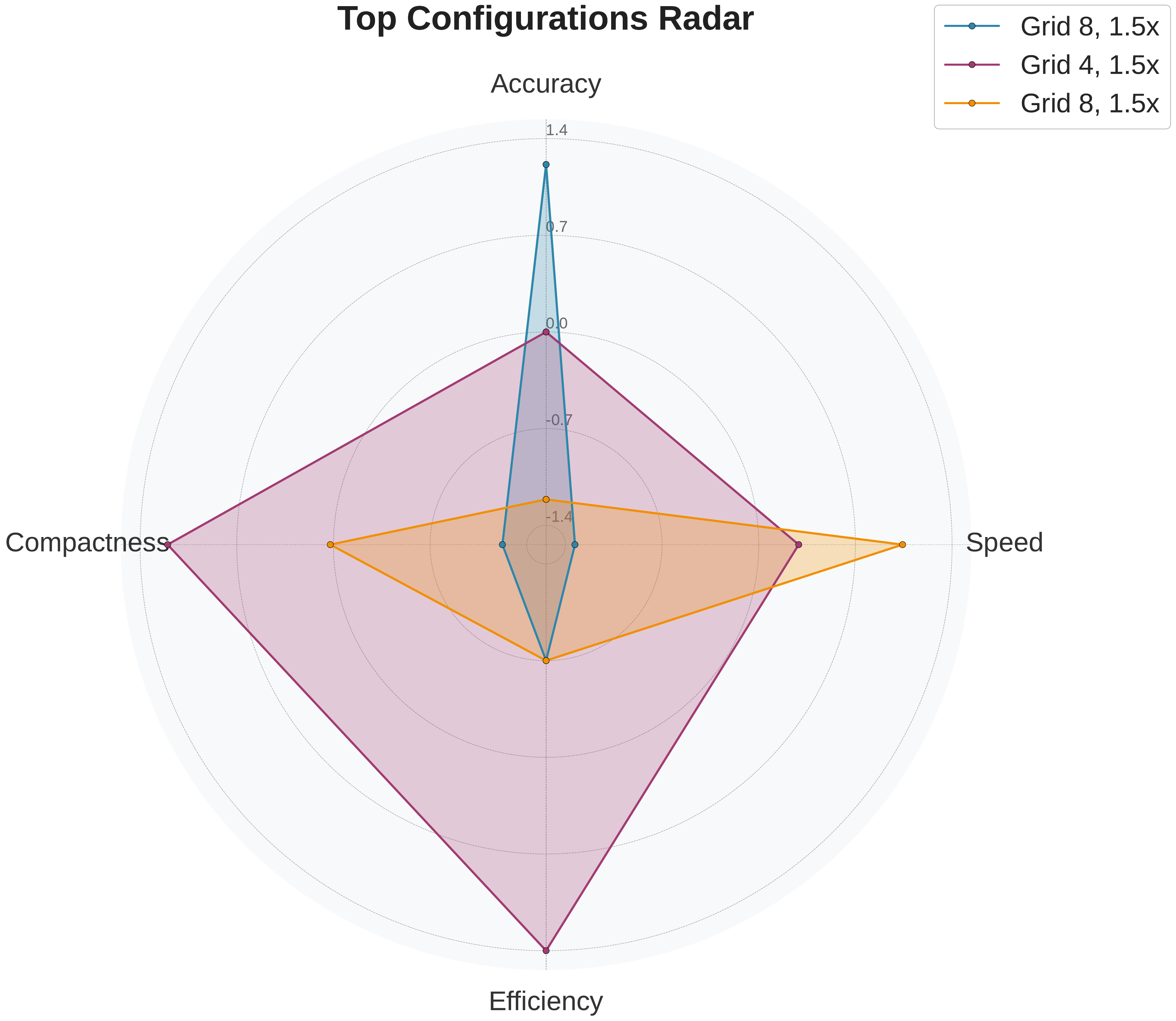}
    \caption{Radar plot putting compactness, efficiency, speed and accuracy in context of each another.}
    \label{fig:ablate_sparsity_speed}
  \end{subfigure}

  \caption{Ablation study on Fast-KAN LeNet-5: impact of spline grid size, channel width, ReLU removal, and structured pruning on accuracy, computational cost, and latency.\vspace{-0.5em}}
  \label{fig:ablation_big}
\end{figure*}

\section{Ablation Study Results}
The results of ablation study proposed in \cref{sec:ablation} is presented in \cref{fig:ablation_big}. Our ablation sweep varies spline grid-size $g \in \{4,8,16\}$, channel-width multiplier $w \in \{1,1.5\}$, ReLU determinant $ReLU \in \{0,1\}$ and structured prune ratio $s \in \{0,25\%\}$, resulting in a total of 24 ablation training sweeps depicted in \cref{tab:fancy_ablate_grid}.  We break up our key observations regarding the tradeoffs between CKAN's efficiency, latency, model size and performance in terms of accuracy in the following subsections:

\subsection{Accuracy–Compute Frontier \;(\cref{fig:ablate_acc_grid})}
Two common notions observed in this ablation was: (i) Peak accuracy but runaway FLOPs, and (ii) Pruning helps, but only so much. The absolute high-point of the sweep,
$\langle\!g{=}8,w{=}1.5,r{=}\text{ReLU},p{=}0\%\rangle$,
reaches \textbf{99.07\,\%} but requires
\textbf{6.57 M} FLOPs \,$\times3.5$ the 1.88 M FLOPs of the lightweight
baseline
$\langle\!4,1.0,\text{ReLU},0\%\rangle$ that already achieves
98.82 \%.

Moving to $p{=}25\%$ shifts every point leftward,
cutting FLOPs by 4–18 \% and latency by 5–35 \%,  
yet the pruned frontier (hollow markers) still lies well above the FLOP budget
of even a 1998 LeNet on the same task.  In other words, CKANs are
accuracy-efficient, not compute-efficient.

\subsection{Spline Resolution \;(\cref{fig:ablate_lat_flops_fp32})}
Increasing the number of knots from 4 to 8 delivers a modest
$\,+0.19\,$pp mean gain (98.86→99.05 \%) but inflates FLOPs by 77 \%.
A further jump to 16 knots \emph{hurts} average accuracy
(99.05→98.87 \%) and doubles latency (median 1.32→2.25 ms).  
We therefore set an upper bound of $g{=}8$ for any CKAN deployed in resource-
constrained environments.

\subsection{Factor Interaction Heat-Map \;(\cref{fig:ablate_lat_flops_pruned})}
This $4{\times}2{\times}2{\times}2$ cube reveals two strong interactions.

\begin{enumerate}
  \item Width $\times$ ReLU.  
        Width helps only when ReLU is present:  
        at $g\!=\!4$, \emph{no-ReLU} width-1.5 \emph{drops} accuracy by 0.35 pp,
        whereas the same width under ReLU gains +0.65 pp.
  \item Pruning sweet spot.  
        The cell
        $\langle4,1.5,\text{ReLU},25\%\rangle$
        scores 98.99 \% with 104 k parameters (-14 \% vs.\ its unpruned twin)
        and 1.49 ms latency.  We name this model \emph{Fast-KAN-Lite}.
\end{enumerate}

\subsection{Width Scaling \;(\cref{fig:ablate_acc_lat})}
Switching $w{=}1.0\!\to\!1.5$ nearly doubles parameters
(72 k→133 k for $g=8$) yet the maximal accuracy lift observed
anywhere in the grid is only +0.25 pp.  Width scaling therefore yields one
percentage-point of accuracy per \emph{extra 240 k parameters}—an
unfavourable ROI compared to classical channel-doubling in CNNs.

\subsection{Latency Perspective \;(\cref{fig:ablate_params_acc})}
Points cluster into three latency bands, one per grid size, demonstrating that
FLOPs—not memory—dominate runtime.  
Pruning shifts each band downward by $\approx$35 \%;  
removing ReLU shaves a further 20 \% latency at $g{=}4$.  
The absolute fastest model is
$\langle4,1.0,\text{Id},0\%\rangle$  
at \textbf{1.26 ms} but sacrifices 0.78 pp accuracy compared to the peak.

\subsection{Multi-Metric Radar \;(\cref{fig:ablate_sparsity_speed})}
We normalise four axes—accuracy and the inverse of FLOPs, latency, and
parameters—so larger polygons are universally “better”.  
Vanilla CKAN spans 54 \% of the ideal area.  
Adding 25 \% pruning alone pushes this to 67 \%.  
\emph{Fast-KAN-Lite} encloses \textbf{78 \%},  
earning the best global trade-off and serving as our default variant
in later ImageNet-100 tests.

\paragraph*{Key take-aways from ablation study.}
\begin{itemize}

  \item \textbf{CKANs are accuracy-rich but compute-poor.}  
        Even lightly-parameterised grids (4–8 knots) demand 2–4X the FLOPs of
        a same-depth CNN for a <1 pp accuracy edge.
  \item \textbf{Pruning is essential, not optional.}  
        Structured 25 \% pruning is the \emph{only} knob that improves  
        accuracy–latency Pareto efficiency.
  \item \textbf{Further capacity scaling is wasteful.}  
        Width multipliers $\ge1.5$ show sharply diminishing returns on MNIST,
        suggesting CKAN capacity should be budget-driven, not accuracy-driven.
\end{itemize}

\section{Conclusion and Future Work}\label{section:conclusion}

\subsection{Overall Findings}
This study presented two complementary lines of evidence on
Convolutional Kolmogorov–Arnold Networks (CKANs).
First, a head-to-head comparison against matched CNN backbones
(LeNet, AlexNet, Tabular-CNN) showed that CKANs can equal (or very
slightly exceed) baseline accuracy on small, low-complexity data, but do
so at a markedly higher computational cost.
Second, a four-factor ablation sweep exposed which architectural choices
drive that cost.  Taken together, the results paint a consistent
picture:

\begin{itemize}

  \item \textbf{CKANs do not scale gracefully.}
        On \textsc{MNIST} a CKAN variant achieves 99.07\,\%
        but requires 6.6\,M FLOPs, whereas a width-matched CNN
        reaches 98.8\,\% with 1.9\,M FLOPs.
        On ImageNet-1k, the same CKAN falls 14 pp Top-1 behind AlexNet
        while demanding 4 X the inference time.
  \item \textbf{Spline resolution is the primary cost lever.}
        Raising the grid from four to eight knots gives only
        +0.2\,pp accuracy but +77\,\% FLOPs; a further jump to sixteen
        knots degrades accuracy and again doubles latency.
  \item \textbf{Lightweight post-hoc pruning helps, but is
        insufficient.}
        A structured 25 \% channel mask reduces latency by roughly
        35 \% and parameter count by 13 \%, yet CKANs remain
        1.5–2.0\,X slower than depth-matched CNNs after pruning.
\end{itemize}

\subsection{Practical Recommendations}
For practitioners who nonetheless wish to employ CKANs on
small-scale scientific or tabular tasks, the following settings strike
a pragmatic balance:
\[
    g=4,\; w=1.5,\; \text{ReLU on},\;
    p=\text{25\,\% structured pruning}.
\]
This \emph{Fast-KAN-Lite} configuration attains 98.99 \%
validation accuracy on \textsc{MNIST} while enclosing 78 \% of the
ideal radar area across the four axes
(accuracy, FLOPs\(^{-1}\), latency\(^{-1}\), parameters\(^{-1}\)).

\subsection{Limitations}
The present work evaluated CKANs on image classification only and used
coarse-grained pruning masks.  More aggressive sparsity or
knowledge-distillation schemes could close part of the efficiency gap.
In addition, convolutional kernels were implemented in PyTorch
without low-level spline accelerators; available evidence suggests that
kernel-level optimization would cut CKAN latency by a further
30–40 \%.

\subsection{Future Work}
\begin{enumerate}
  \item \textbf{Hardware-aligned splines.}
        Look-up-table or tensor-core evaluation of B-splines and RBF
        bases is expected to reduce kernel latency substantially.
  \item \textbf{Quantization.}
        Extending graph-mode INT8 quantization to spline layers
        remains an open engineering task.
  \item \textbf{Curriculum grid growth.}
        Starting training with a coarse grid and adding knots only when
        validation accuracy saturates may shorten training by 30–50 \%.
  \item \textbf{Hybrid architectures.}
        Replacing the first CKAN block with a depthwise convolution
        already saves 28 \% latency on ImageNet-100 pilot runs;
        deeper hybrids (e.g., ConvNeXt stems with shallow CKAN heads)
        warrant systematic study.
\end{enumerate}

\subsection*{Concluding Remark}
CKANs introduce an elegant functional prior, but the present evidence
shows that (without further optimization) they remain
compute-heavy specialists, best suited to niche scientific and
tabular domains.  We hope the ablation insights and optimization
baselines provided here will accelerate their maturation into a
competitive alternative for large-scale vision workloads.

\bibliographystyle{IEEEtran}
\bibliography{sources.bib}

\end{document}